\documentclass[sigconf,anonymous=false,review=false]{acmart}

\AtBeginDocument{%
  }

\copyrightyear{2022}
\acmYear{2022}
\setcopyright{acmlicensed}\acmConference[KDD '22]{Proceedings of the 28th ACM SIGKDD Conference on Knowledge Discovery and Data Mining}{August 14--18, 2022}{Washington, DC, USA}
\acmBooktitle{Proceedings of the 28th ACM SIGKDD Conference on Knowledge Discovery and Data Mining (KDD '22), August 14--18, 2022, Washington, DC, USA}
\acmPrice{15.00}
\acmDOI{10.1145/3534678.3542674}
\acmISBN{978-1-4503-9385-0/22/08}

\usepackage{bibentry}
\usepackage{amsfonts}
\usepackage{multirow}
\usepackage{graphics}
\usepackage{paralist}
\usepackage{makecell}
\usepackage{natbib}
\usepackage[capitalise]{cleveref}
\usepackage{stfloats}
\usepackage{algorithmic}
\usepackage{amsmath}
\usepackage{amsthm}
\usepackage[ruled]{algorithm2e}
\begin{document}

\title{Medical Dialogue Response Generation with Pivotal Information Recalling}

\newcommand\ModelName{MedPIR}





\author{Yu Zhao}
\authornote{Both authors contributed equally to this research.}
\email{zhaoyuhitsz@163.com}
\affiliation{%
\institution{Harbin Institute of Technology}
  \city{Shenzhen}
  \country{China}
}

\author{Yunxin Li}
\authornotemark[1]
\email{liyunxin987@163.com}
\affiliation{%
\institution{Harbin Institute of Technology}
  \city{Shenzhen}
  \country{China}
}

\author{Yuxiang Wu}
  \email{yuxiang.wu@cs.ucl.ac.uk}
\affiliation{%
\institution{University College London}
  \city{London}
  \country{United Kingdom}
}

\author{Baotian Hu}
\email{hubaotian@hit.edu.cn}
\authornote{Corresponding author.}
\affiliation{%
\institution{Harbin Institute of Technology}
  \city{Shenzhen}
  \country{China}
}

\author{Qingcai Chen}
\email{qingcai.chen@hit.edu.cn}
\affiliation{%
\institution{Harbin Institute of Technology}
  \city{Shenzhen}
  \country{China}
}

\author{Xiaolong Wang}
\email{xlwangsz@hit.edu.cn}
\affiliation{%
\institution{Harbin Institute of Technology}
  \city{Shenzhen}
  \country{China}
}

\author{Yuxin Ding}
  \email{yxding@hit.edu.cn}
\affiliation{%
\institution{Harbin Institute of Technology}
  \city{Shenzhen}
  \country{China}
}

\author{Min Zhang}
\email{zhangmin2021@hit.edu.cn}
\affiliation{%
\institution{Harbin Institute of Technology}
  \city{Shenzhen}
  \country{China}
}

\renewcommand{\shortauthors}{Yu Zhao and Yunxin Li, et al.}

\begin{abstract}
Medical dialogue generation is an important yet challenging task.
Most previous works rely on the attention mechanism and large-scale pretrained language models. 
However, these methods often fail to acquire pivotal information from the long dialogue history to yield an accurate and informative response, 
due to the fact that the medical entities usually scatters throughout multiple utterances along with the complex relationships between them.
To mitigate this problem, we propose a medical response generation model with \emph{Pivotal Information Recalling}~(\ModelName), which is built on two components, i.e., knowledge-aware dialogue graph encoder and recall-enhanced generator.
The knowledge-aware dialogue graph encoder constructs a dialogue graph by exploiting the knowledge relationships between entities in the utterances, and encodes it with a graph attention network.
Then, the recall-enhanced generator strengthens the usage of these pivotal information by generating a summary of the dialogue before producing the actual response. 
Experimental results on two large-scale medical dialogue datasets show that \ModelName{} outperforms the strong baselines in BLEU scores and medical entities F1 measure. 
\end{abstract}

\ccsdesc[500]{Applied computing~Health care information systems}
\ccsdesc[500]{Computing methodologies~Discourse, dialogue and pragmatics}
\ccsdesc[500]{Computing methodologies~Natural language generation}

\keywords{medical dialogue generation, pivotal information recalling}

\maketitle

\section{Introduction}

\begin{figure}[t]
\centering
\includegraphics[scale=0.67]{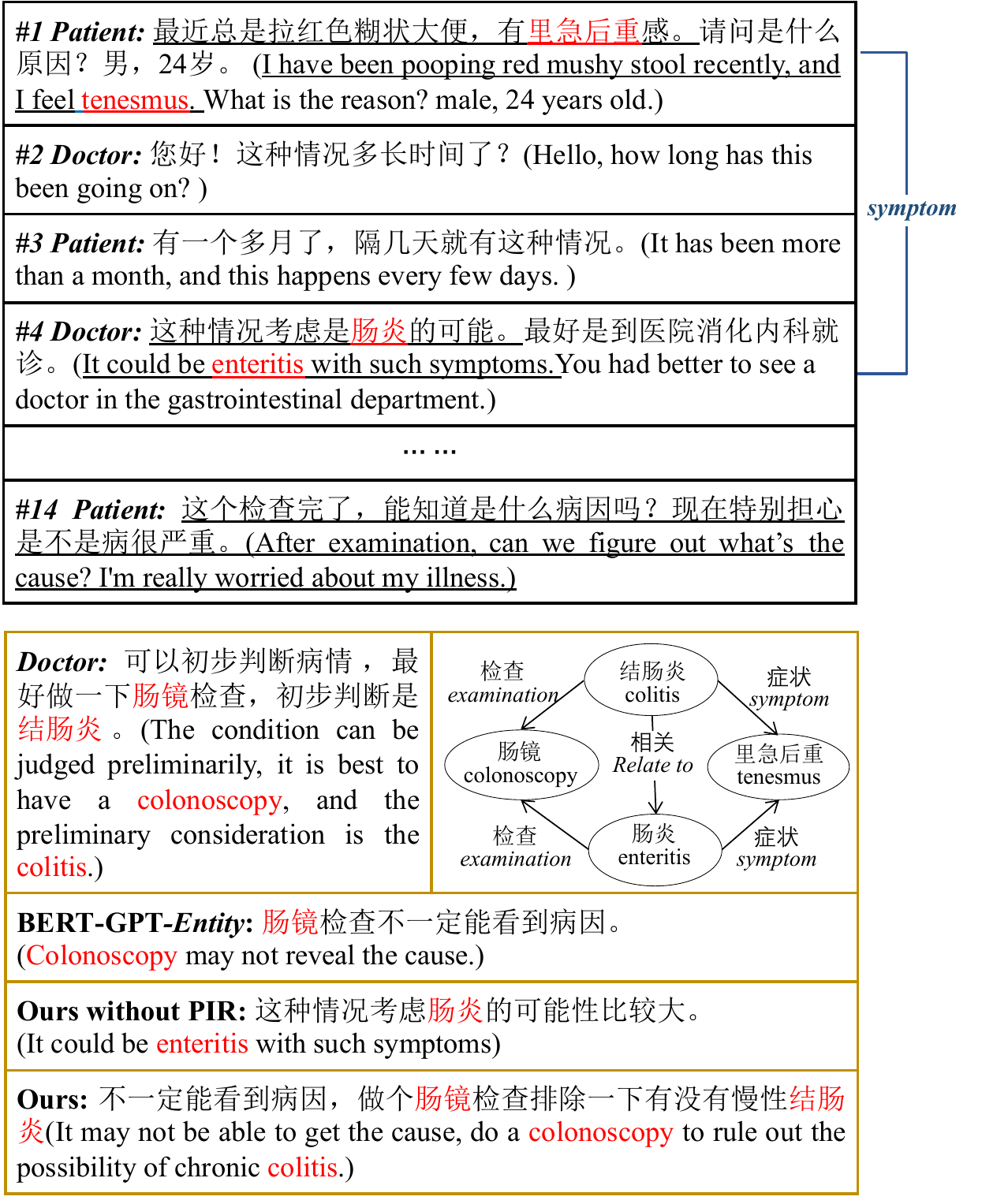}
\caption{An excerpted medical dialogue from MedDG~\citep{meddg}. The colored words are key medical phrases and the underlined parts represent the pivotal information to induce the response. The knowledge graph shown on the right is usefull for  diagnosing. The responses generated by the baselines and our proposed method are shown at the bottom.} 
\label{fig:conversation}
\end{figure}

Medical dialogue system (MDS) has received much attention due to its high practical value. Previous works~\citep{bertgpt, meddg, vrbot} usually model the dialogue history as sequential text and employ the sequence-to-sequence~(Seq2Seq) models that built on large-scale pretrained text encoder and decoder to generate medical responses.  

To have a comprehensive understanding of the patient, medical dialogues are always relatively long, and there are rich medical terminologies scattered in multiple utterances. Some works~\citep{madotto2020learning, graph-evolving, enhance-graph, learn-to-infer-entity}
introduce the external medical knowledge into the Seq2Seq models and show that it can improve the performance. But these works fall short in utilizing the complex medical relationships between different utterances, which is important for inducing the next response. As shown in Figure~\ref{fig:conversation}, the entities \textit{tenesmus} and \textit{enteritis} indicate the \textit{symptom} relationship between utterance\textit{\#1} and utterance\textit{\#4}. Due to ignoring the medical relationship between utterances, the strong baseline model BERT-GPT-\emph{Entity}~\citep{bertgpt} misses the pivotal entity \textit{colitis} in the generated response. Our MedPIR derives the \textit{colitis} from \textit{enteritis} and generate a more accurate response. 

How to acquire pivotal information from long dialogue history is the core of MDS. Previous works heavily rely on the cross-attention mechanism to use dialogue history, which falls short in locating the key information from a long sequence. This issue may be caused by the fact that the cross-attention mechanism is not trained with explicit supervision signals when recalling pivotal information. Recent works~\citep{vrbot,med_dialogue_smmary,extract_symptoms,mie} proposed to extract the medical key phrases and sentences from the dialogue history and incorporate them into response generation via the cross-attention mechanism as well. However, these works bypass the fundamental problem of utilizing medical relations between different utterances, and fail to fully exploit the pivotal information from dialogue history during response generation. 

The above investigation suggest that it is important to model the complex medical relationships between multiple utterances and explicitly guide the decoder to make full use of the pivotal information during response generation. In this work, we propose a Medical response generation model with Pivotal Information Recalling (\ModelName), where we enforce the generator to recall pivotal information during generation. 
It mainly contains the knowledge-aware dialogue graph encoder and recall-enhanced generator. 

The knowledge-aware dialogue graph encoder exploits the knowledge relationship between medical entities scattered in different utterances to construct the dialogue graph. And its representation acquired with graph attention networks is fed to the generator. Hence, the knowledge-aware dialogue graph encoder can facilitate the generator to use pivotal medical information distributed in multiple utterances from the perspective of the global dialogue structure. The recall-enhanced generator is designed to explicitly generate the pivotal information from long dialogue history first. And then, the pivotal information sequence is used as the prefix of response to prompt to generate more focused responses. In this way, the recall-generator enforces the cross-attention mechanism to fully use the pivotal information from the encoder with the recall signal. Moreover, the recall-enhanced generator also strengthens the interaction between the response and pivotal information recalled from dialogue history via the self-attention mechanism within the decoder. Besides, we also retrieve relevant knowledge from the medical knowledge graph CMeKG~\citep{cmekg} and use the medical pre-trained model to obtain an in-depth understanding of medical knowledge. 

Our contributions can be summarized as follows:

    \textit{1)} We propose an MDS model with pivotal information recalling~(\ModelName). It can exploit the complex medical relationship between dialogue utterances via the knowledge-aware dialogue graph encoder and recall pivotal information from long dialogue history to produce accurate responses in the recall-enhanced generator. 
    
    \textit{2)} We conduct extensive experiments on large-scale medical dialogue datasets MedDG~\citep{meddg} and MedDialog~\citep{bertgpt}.The experimental results show that our proposed model achieves new state-of-the-art results by outperforming previous strong baselines VRBot~\citep{vrbot} and BERT-GPT-\emph{Entity}~\citep{meddg} on BLEU and medical entities F1 metrics.



\section{Related Works}

\quad \textit{Medical Dialogue System (MDS).} Previous MDS works mostly adopt a sequence-to-sequence framework~\citep{attention, transformer}. It consists of a context encoder to encode the dialogue history and a decoder to generate the response. Since the medical dialogue is often long and contains professional medical information, it is difficult for the attention mechanism to attend on the pivotal information in the dialogue history. To recognition key information in medical dialogues, \citet{extract_symptoms} and \citet{mie} 
extract patient's symptoms and medical status from history. Most recent, \citet{vrbot} proposed a variational medical dialogue generation model strengthens by summarizing diagnosis history through a key phrase. However, these method only extract key information by phrases and cannot make fully use of the complicated pivotal information scattered in dialogue history. Different from previous works, we build medical dialogue graph that exploits medical relationship between utterances, and train the model to generate the pivotal information before producing the actual response, so that the model can learn to focus on the key information.

\textit{Dialogue Graph Construction.} To model the relationship between utterances in a dialogue,~\citet{xu-etal-2021-discovering, chen-etal-2018-structured, graph} propose to construct a dialogue structure graph based on dialogue state transitions. \citet{ddams} proposed to model the dialogue structure of the meeting by modeling different discourse relations. However, they did not exploit external knowledge base, which is essential for producing medical dialogue response. In contrast, we construct a knowledge-aware dialogue graph by incorporating external medical knowledge from CMeKG.

\textit{Knowledge-grounded Dialogue Generation.} Recent works~\citep{li2020zero, chen2020bridging, ghazvininejad2018knowledge} proposed to improve the performance of dialogue modeling by retrieving relevant knowledge from the commonsense graph, such as ConceptNet~\citep{conceptnet}, and incorporating the object facts in generation. \citet{learningtoselect, low-resource-kg, sequential-kg, wizardofwiki} facilitated knowledge-ground dialogue generation by retrieving from unstructured documents. \citet{vrbot} and \citet{graph-evolving} used medical knowledge graph to guide response generation through copy mechanism \cite{pinternet}, but they did not use medical knowledge graph to exploit dialogue structure. In this work, the external knowledge is used to construct dialogue graph and is also encoded with a knowledge encoder.

\section{Methodology}
The key information of medical dialogue often scatters throughout the long history, making it difficult for traditional MDS models to acquire pivotal information from the dialogue history. In this section, we first describe the base medical response generation model in~\cref{sec:baseline}. Then, we introduce two techniques to improve the recalling of pivotal information from the dialogue -- knowledge-aware dialogue graph encoder (\cref{sec:graph_enc}) and recall-enhanced generator (\cref{sec:recgen}). Finally, the training method of our proposed method is presented in~\cref{sec:training}.

\subsection{Base Model} \label{sec:baseline}
Most previous works in dialogue response generation~\citep{bertgpt, meddg} adopt the sequence-to-sequence architecture to model the dialogue history and exploit external medical knowledge~\citep{vrbot,graph-evolving, enhance-graph} to generate the response. For our base model, we follow~\citet{bertgpt} and use BERT-GPT as the backbone of our encoder and the generator. Given a dialogue history between a doctor and a patient $X=( X_1, X_2,...,X_M )$, where $X_i = (x_{i,1}, x_{i,2}, ... x_{i,|X_i|} )$ is $i$-th utterance in the dialogue history with $|X_i|$ tokens, the context encoder encodes the concatenation of utterances to obtain the context encoding $\mathbf{H}_{ctx}$.

We also follow previous works~\citep{wizardofwiki, vrbot, mie} to retrieve external knowledge and use a knowledge encoder to obtain the knowledge encoding $\mathbf{H}_{k}$ (more details are elaborated in \cref{sec:filter}). The base model produces responses $Y = (y_1, y_2,...,y_{|Y|})$ conditioned on both $\mathbf{H}_{ctx}$ and $\mathbf{H}_{k}$.

\subsection{Knowledge-aware Dialogue Graph Encoder}
\label{sec:graph_enc}
Since the base dialogue model only views the medical dialogue history as a sequence of utterances, it is hard to model the diverse medical causal relationships between different utterances~\citep{ddams}, which implies the pivotal medical information for inducing the next response. To tackle this problem, we propose the Knowledge-aware Dialogue Graph Encoder (KDGE) that constructs a dialogue graph with knowledge, and then encodes the graph with a graph attention network.

\begin{figure}[t]
    \centering
    \includegraphics[scale=0.85]{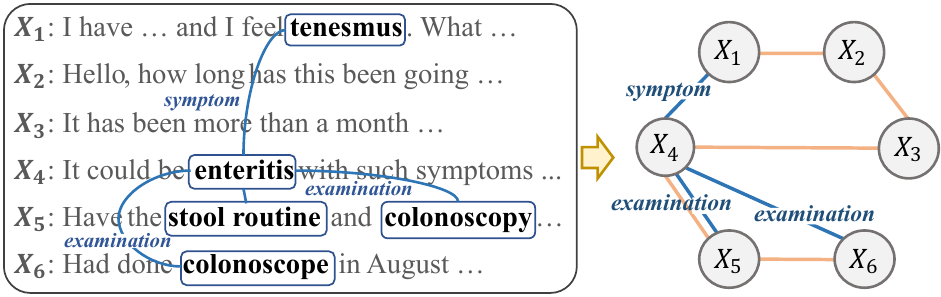}
    \caption{
    A part of medical dialogue and the corresponding dialogue graph we construct. The \textit{blue edges} connect the utterances with medical relations revealed by medical entities, the orange edges connect the neighbouring utterances. 
    }
    \label{fig:graph}
\end{figure}

First, we transform the sequential dialogue history into a graph. Each utterance is regarded as a vertex, and there are two types of edge between the vertices. One type of edge connects the neighboring utterances, which denotes the normal temporal relations like previous works~\citep{xu-etal-2021-discovering, chen-etal-2018-structured}. The other type is \emph{knowledge-aware edge}, which connects the scattered utterances with medical relationships. 
These knowledge-aware edges incorporates medical knowledge from external medical knowledge graph into the dialogues, allowing the model to represent complex medical relationships of the utterances. More concretely, we first extract medical entities from each utterance, and then look up the relationships between them from an external knowledge graph.\footnote{We choose CMeKG as our medical knowledge graph because it is the largest Chinese medical knowledge graph that is publically available.} We add a knowledge-aware edge between two utterances if there exists a relationship between the medical entities from the two utterances. \cref{fig:graph} shows an example of this construction process. In the left part, the bold words are entities scattered in utterances, and the blue lines connect entities with certain relations. The right part represents the constructed knowledge-aware dialogue graph.

With the constructed knowledge-aware dialogue graph $G$, we then apply Relational Graph Attention Network (RGAT) proposed by~\citet{rgat} to encode these pivotal relational information in the dialogue.
For each vertex $v_i$ in $G$, we use a transformer-based encoder to encode its corresponding utterance, and compute the average of the token representations as the utterance embedding. Then the utterance embedding is concatenated with its speaker embedding (a trainable embedding that represents the role of the speaker) to form $v_i$'s initial vertex embedding $\mathbf{v}_i^0$. At last, RGAT is used to compute the updated encoding of the vertices:
\begin{equation}
 (\mathbf{v}_1, ..., \mathbf{v}_M )  =  RGAT \left( (\mathbf{v}_1^0, ..., \mathbf{v}_M^0), G\right).
\end{equation}

To perform dialogue recalling, we regard the context encoding as initial history representation, and define recall score $\alpha_{v_i}$ as the importance of utterance $X_i$ for recalling as follows:
\begin{equation}
    \label{eq:vertex_alpha}
    \alpha_{v_i} = \sigma\left((\mathbf{W}_v^q \mathbf{h}_{ctx})^T (\mathbf{W}_v^k \mathbf{v}_i)\right),
\end{equation}
where $\mathbf{h}_{ctx}$ is mean-pooled from $\mathbf{H}_{ctx}$,
$\mathbf{W}_v^q$ and $\mathbf{W}_v^k$ are trainable parameters,
$\sigma$ denotes the sigmoid function.
Then the final structure encoding of $X_i$ is obtained from the addition of utterance encoding $\mathbf{h}_{i}$ and vertex encoding $\mathbf{v}_i$ weighted by the corresponding recall score:
\begin{equation}
    \mathbf{h}_{stc,i} = \alpha_{v_i} (\mathbf{h}_{i} + \mathbf{v}_i).
\end{equation}
The concatenation of $\{\mathbf{h}_{stc,i}\}_{i=1}^{M}$ is the final structure encoding, denoted as $\mathbf{H}_{stc}$.

\begin{figure*}[t]
    \centering
    \includegraphics[scale=0.9]{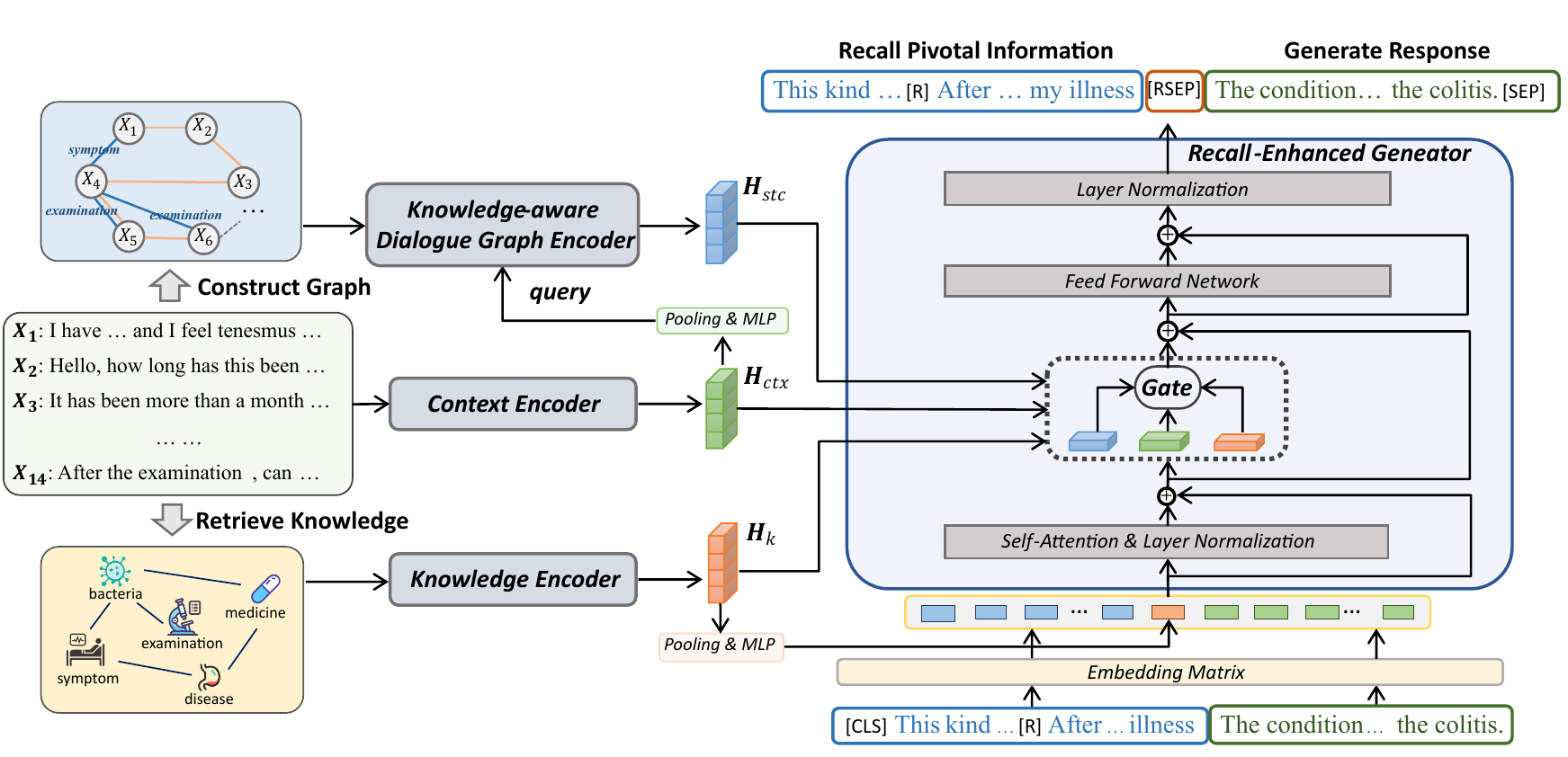}
    \caption{The overall architecture of \ModelName. The context encoder encodes dialogue history into context encoding first. Then, the knowledge-aware dialogue graph encoder encodes dialogue graph and uses context encoding as the query to obtain the final structure encoding. The knowledge encoder encodes the retrieved medical knowledge from CMeKG. The right part shows the recall-enhanced generator. }
    \label{fig:overall}
\end{figure*}

\subsection{Recall-Enhanced Generator}
\label{sec:recgen}

In the base model, the generator first performs unidirectional self-attention with the generated sequence to obtain the decoding state at each time, and then exploits $\mathbf{H}_{ctx}$ and $\mathbf{H}_k$ by the cross-attention mechanism.
When this dialogue model is only trained to produce the response, its attention mechanism is often overwhelmed with the long dialogue history and fails to focus on the pivotal information. 
We propose \emph{Recall-Enhanced Generator} (REG) to explicitly generate the pivotal information $\mathcal{R}$ before producing the response. $\mathcal{R}$ is a brief summary that contains key medical information of the dialogue history. After producing $\mathcal{R}$, it will continue to generate focused response as follows:
\begin{equation}
y_t = REG(\mathbf{H}_{ctx}, \mathbf{H}_k, \mathbf{H}_{stc},[\mathcal{R};y_{<t}]) ,
\end{equation}

At training time, $\mathcal{R}$ is automatically constructed with medical pretrained model PCL-MedBERT (more details introduced in Section~\ref{sec:training}) to serve as a supervision signal to train the model to recall pivotal information.
At test time, \ModelName{} will first produce the recalled information and then generate the response. There are two main advantages of the method: 
\textit{1)}~the qualified pre-generated recall $\mathcal{R}$ provides a shortcut for the generator to access key history information through self-attention;
\textit{2)}~recalling strengthens the cross-attention mechanism to attend to the pivotal information provided by the encoders.

As shown in the right half of \cref{fig:overall}, tokens are first converted to embedding through the embedding matrix as the initial hidden state inputting to the generator. 
Then, REG sequentially generates the recalled pivotal information $\mathcal{R}$, a separator, and finally the target response $Y$.
Note that we use the average pooled knowledge encoding as the embedding of separator to drive the knowledge fusion during generation, as shown in the bottom-right part.

More specifically, REG consists of multiple layers decoder block. Let $\mathbf{h}^{l-1}_{t}$ denote the output of ($l-1$)-th layer at $t$ step. The calculating process in $l$-th block can be formulated as:
\begin{equation}
\mathbf{h}^l_{S,t} = 
\textit{LayerNorm}\left(
\textit{SA}(\mathbf{h}^{l-1}_t) + \mathbf{h}^{l-1}_t
\right),
\end{equation}
\begin{equation}
    \mathbf{h}^l_{F,t} = Fusion\left(
        \mathbf{H}_{ctx}, \mathbf{H}_{stc}, \mathbf{H}_{k}
    \right) +  \mathbf{h}^l_{S,t},
\end{equation}
\begin{equation}
\mathbf{h}^l_t=\textit{LayerNorm} \left(\textit{FFN} 
 (\mathbf{h}^l_{F,t})
 +  \mathbf{h}^l_{F,t}
\right),
\end{equation}
where $\textit{SA}$ denotes unidirectional self-attention in decoder, and $\textit{FFN}$ is a feed-forward network.

To integrate different type of information from the encoders, we introduce the $\textit{Fusion}(\text{·})$ operation, a gating mechanism that combines
the context encoding $\mathbf{H}_{ctx}$, structure encoding $\mathbf{H}_{stc}$, and knowledge encoding $\mathbf{H}_{k}$.
It first condenses multifaceted encoding by taking $\mathbf{h}_{S,t}^l$ as the query to perform cross-attention ($\textit{CA}$) with $\mathbf{H}_{ctx}$, $\mathbf{H}_{stc}$ and $\mathbf{H}_{k}$ respectively, and then conduct weighted sum of the condensed encodings with the gate scores:
\begin{equation}
\label{eq:fusion}
\begin{split}
Fusion\left(\text{·} \right)=& 
\;g_{ctx}^l\textit{CA}^l (\mathbf{H}_{ctx}, \mathbf{h}_{S,t}^l) + g_{k}^l\textit{CA}^l (\mathbf{H}_{k}, \mathbf{h}_{S,t}^l) \\
\;&+ g_{stc}^l\textit{CA}^l (\mathbf{H}_{stc}, \mathbf{h}_{S,t}^l),
\end{split}
\end{equation} 
where the gate scores $g_{ctx}$, $g_{stc}$ and $g_{k}$ are obtained by a linear layer with sigmoid function:
\begin{equation}
g^{l}= \sigma\left(\mathbf{W}^l\textit{CA}^l (\mathbf{H} , \mathbf{h}_{S,t}^l )\right) .
\end{equation}
Then, the three gate scores are normalized by the softmax function to obtain the final gate scores applied in \cref{eq:fusion}.

At the last layer, an output projection layer is applied to get the final generating distribution $p_t$ over vocabulary:
\begin{equation}
\label{eq:output_layer}
p_t = softmax\left(\mathbf{W}_v \mathbf{h}_{t}^L +b_v  \right) .
\end{equation}

While recalling pivotal information and generating response, the gate-based fusion network dynamically controls the inflows of context encoding, structure encoding, and knowledge encoding.
The structure encoding obtained from KDGE provides complementary information to the context encoding, facilitating REG to recall pivotal information. This behavior can be demonstrated by the visualization of the gate scores in the \cref{fig:gatescore}.

\subsection{Training}

\label{sec:training}
\subsubsection{Recall Supervision Signals}
\label{sec:supervised_signals}

The ideal recall sequence $\mathcal{R}$ is a summary of the current dialogue. But medical dialogue summary is not annotated in most cases. To deal with this problem, we introduce PCL-MedBERT to select the utterances that are most relevant to the target response as training signals. First, PCL-MedBERT encodes $X_i$ and $Y$ into $\mathbf{h}_{i}^r$ and $\mathbf{h}_{y}^r$ respectively, and we use the cosine-similarity between them to score $X_i$:
\begin{equation}
    sim(X_i, Y) = \frac{\mathbf{h}_{i}^r \cdot \mathbf{h}_{y}^r}{\Vert \mathbf{h}_{i}^r \Vert \Vert \mathbf{h}_{y}^r \Vert} .
\end{equation}
Then, we select $k$ utterances with highest similarity scores, denoted as $X^r = (X_{1}^r...X_{k}^r)$. The concatenation of $X^r$ is used as the target recall $\mathcal{R}$ for training recall generation. Despite that this is a distantly-supervised method, the utterances extracted by PCL-MedBERT\footnote{https://code.ihub.org.cn/projects/1775}  usually contain pivotal information for generating an informative medical response (see \cref{fig:case} for an example of extracted and generated recall sequence).
To further facilitate the model to generate qualify $\mathcal{R}$ at inference, we also train it to identify pivotal utterances by supervising the recall score $\alpha_{v_i}$ (obtained by \cref{eq:vertex_alpha}) through binary cross-entropy:
\begin{equation}
    \mathcal{L}_r  =  \sum_{i=1}^{M} -r_i \log \alpha_{v_i} - (1-r_i) \log ( 1-\alpha_{v_i} ) ,
\end{equation}
where $r_i \in \{0,1\}$ indicates whether $X_i$ is in $X^r$. The higher $\alpha_{v_i}$, the more important $X_i$ is for recalling.

\subsubsection{Overall Training Objective}
We minimize the negative log-likelihood of the recall sequence $\mathcal{R}=(s_1, s_2, ...,s_{|\mathcal{R}|})$ and response $Y$, where $Y$ is generated after $\mathcal{R}$:
\begin{equation}
\mathcal{L}_{\mathcal{R}} = \sum_{i=1}^{|\mathcal{R}|} - \log p(s_i | X, s_{<i}) ,
\end{equation}
\begin{equation}
\mathcal{L}_Y = \sum_{i=1}^{|Y|} - \log p(y_i | X, \mathcal{R}, y_{<i}) .
\end{equation}
Then we jointly optimize $\mathcal{L}_Y$, $\mathcal{L}_{\mathcal{R}}$ and $\mathcal{L}_r$ weighted by $\lambda_Y$, $\lambda_{\mathcal{R}}$ and $\lambda_r$,  respectively:
\begin{equation}
\mathcal{L}  =  \lambda_Y \mathcal{L}_Y + \lambda_\mathcal{R} \mathcal{L}_\mathcal{R} + \lambda_r \mathcal{L}_r.
\end{equation}
We present the overall training algorithm in Algorithm~(\ref{algo:trainning}).

\begin{algorithm}[h]
\caption{Training Algorithm}
\label{algo:trainning}
\LinesNumbered
\KwIn{training dialogue dataset $\mathcal{D}$, initial parameter of MedPIR $\theta$, learning rate $\gamma$, PCL-MedBERT}

\While{not converged}{
    \ForEach{sample $(X, y)$ in $\mathcal{D}$}{
        Obtain $X^r$ by PCL-MedBERT;
        
        Calculate $\{\alpha_{v_i}\}_{i=1}^{M}$ by Eq.(\ref{eq:vertex_alpha});
        
        $\mathcal{L}_r  \gets  \sum_{i=1}^{M} -r_i \log \alpha_{v_i} - (1-r_i) \log ( 1-\alpha_{v_i} )$ ;

        Calculate $p(s_i | X, s_{<i})$ and $p(y_i | X, \mathcal{R}, y_{<i})$ by Eq.(\ref{eq:output_layer}) 
        
        $\mathcal{L}_{\mathcal{R}}  \gets  \sum_{i=1}^{|\mathcal{R}|} - \log p(s_i | X, s_{<i})$ ;
    
        $\mathcal{L}_Y  \gets  \sum_{i=1}^{|Y|} - \log p(y_i | X, \mathcal{R}, y_{<i})$ ;
    
        $\mathcal{L}  \gets   \lambda_Y \mathcal{L}_Y + \lambda_\mathcal{R} \mathcal{L}_R + \lambda_r \mathcal{L}_r$ ;
        
        $\theta \gets \theta - \gamma \nabla \mathcal{L}$;
    }
}
\end{algorithm}
 
\section{Experiments}
\begin{table*}[h]
\renewcommand\arraystretch{1.2} 
\centering
    \begin{tabular}{@{\quad}lcccc|cccccc@{\quad}}
        \hline  
         \multirow{2}{*}{\textbf{Model}} & \multicolumn{4}{c|}{\textbf{Sequence Metrics}} & \multicolumn{6}{c}{\textbf{Entity Metrics}} \\ \cline{2-11} 
        & \textbf{B@1} & \textbf{B@2} & \textbf{B@4} & \textbf{D@2}  &  \textbf{F1} & \textbf{F1-D} & \textbf{F1-S} & \textbf{F1-A} & \textbf{F1-T} & \textbf{F1-M} \\ \hline
        Seq2Seq~\citep{seq}               & 0.3852  & 0.3487  & 0.3297  & 0.8561  & 0.113 & 0.096 & 0.068 & 0.395 & 0.096 & 0.055\\ \hline
        Seq2Seq-\emph{Entity}~\citep{meddg}  & 0.3884  & 0.3416  & 0.3380  & 0.8635  & 0.195 & 0.224 & 0.159 & 0.406 & 0.178 & 0.107 \\ \hline
        HRED~\citep{hred}                   & 0.3819  & 0.3365  & 0.3345  & 0.8670  & 0.109 & 0.097 & 0.064 & 0.383 & 0.098 & 0.053 \\ \hline
        HRED-\emph{Entity}~\citep{meddg}    & 0.3942  & 0.3386  & 0.3255  & 0.8731  & 0.195 & 0.232 & 0.155 & 0.411 & 0.191 & 0.106 \\ \hline
        DialoGPT~\citep{dialogpt}               & 0.3122  & 0.3125  & 0.3266  & 0.7869  & 0.122 & 0.100 & 0.089 & 0.409 & 0.104 & 0.094 \\ \hline
        DialoGPT-\emph{Entity}~\citep{meddg} & 0.3193  & 0.3106  & 0.3446  & 0.7892  & 0.176 & 0.180 & 0.095 & 0.366 & 0.203 & 0.094 \\ \hline

        BERT-GPT~\citep{bertgpt}               & 0.4260  & 0.3593  & 0.3344  & 0.8893  & 0.146 & 0.138 & 0.099 & 0.399 & 0.106 & 0.101 \\ \hline
        BERT-GPT-\emph{Entity}~\citep{meddg} & 0.4286  & 0.3545  & 0.3187  & 0.8976  & 0.207 & 0.236 & 0.171 & 0.410 & 0.208 & 0.131 \\ \hline
        VRBot~\citep{vrbot}                  & 0.3455  & 0.3144  & 0.3306  & 0.7460       & 0.075 & 0.073     & 0.052     & 0.194     & 0.100     & 0.035     \\ \hline
        \hline
        \ModelName{} (Ours) & \textbf{0.4476} & \textbf{0.3866} & \textbf{0.3621} & 0.8915 & \textbf{0.227} & \textbf{0.263} & \textbf{0.175} & \textbf{0.413} & \textbf{0.213} & \textbf{0.144} \\\hline
        \hspace{0.4em} $-$ \footnotesize{Knowledge-aware dialogue graph encoder (KDGE)} & 0.4109  & 0.3317  & 0.2888  & \textbf{0.8976}  & 0.216 & 0.258 & 0.170 & 0.413 & 0.212 & 0.135  \\ \hline
        \hspace{0.4em} $-$ \footnotesize{Recall-enhanced generator (REG)}  & 0.4247  & 0.3541  & 0.3353  & 0.8897  & 0.220 & 0.262 & 0.175 & 0.407 & 0.210 & 0.141 \\ \hline
        \hspace{0.4em} $-$ \footnotesize{Knowledge encoder} & 0.4379  & 0.3738  & 0.3573  & 0.8848  & 0.144 & 0.150 & 0.095 & 0.385 & 0.137 & 0.082 \\ \hline
        \hspace{0.4em} $-$ \footnotesize{KDGE \& REG}  & 0.4023  & 0.3308  & 0.2964  & 0.8946  & 0.220 & 0.260 & 0.175 & 0.412 & 0.212 & 0.139 \\ \hline

    \end{tabular}

    \caption{Automatic evaluation results on MedDG dataset. The models with `-\emph{Entity}' suffix denotes their inputs incorporate entities by concatenating them with history directly. The entity F1 scores of different categories: F1-D (Disease), F1-S (Symptom), F1-A (Attribute), F1-T (Test) and F1-M (Medicine). B@n denotes BLEU-n and D@2 denotes DISTINCT-2.  }
    \label{tab:expresult}
\end{table*}

\subsection{Settings}

\subsubsection{Datasets}
We adopt two medical dialogue datasets MedDG~\citep{meddg} and MedDialog~\citep{bertgpt} to evaluate our proposed model. 
Both of them are collected from online consultation websites.
In MedDG, the training/development/test sets contain $14864$/$2000$/$1000$ dialogues respectively, where each utterance is semi-automatically annotated with $5$ types with a total of $160$ medical entities.
\citet{vrbot} pointed that most dialogues in MedDialog have less than $5$ utterances, which also contain few medical professional information. Thus, we follow the refined version of MedDialog preprocessed by~\citet{vrbot} to evaluate our method, where the training/development/test sets include $32723$/$3000$/$3000$ dialogues respectively. 
\subsubsection{Evaluation Metrics}
We use BLEU~\citep{bleu} to evaluate the n-gram lexical similarity, and use DISTINCT~\citep{distinct} to evaluate the diversity of the generated responses.
We also take medical entities F1 score as an important metric, which can better evaluate the actuality of medical response than lexical similarity metrics.
In MedDG dataset, we use the published script\footnote{https://github.com/lwgkzl/MedDG} to recognize entities in responses,
and evaluate different types of entity respectively.
Due to MedDialog is not annotated with entities, we first collect medical entities in CMeKG, then extract entities in responses by string matching with the collected entities.
Besides, we conduct human evaluation to evaluate the responses' fluency, coherence, and correctness. The fluency only measures whether the generated response is fluency, while coherence measures whether the generated response is smooth and logical with context. The correctness evaluates whether the responses uses correct medical knowledge. Three metrics are scored by annotators with a range from 1 (bad) to 5 (excellent).

\subsubsection{Baselines}
We use Seq2Seq~\citep{seq} and HRED~\citep{hred} as RNN-based dialogue generation baselines. Compared to Seq2Seq, HRED uses hierarchical encoders to model the dialogue context from token level and utterance level. DialoGPT~\citep{dialogpt} and BERT-GPT~\citep{bertgpt} are transformers-based pre-trained dialogue response models. DialoGPT is pre-trained on open-domain dialogue corpora, while BERT-GPT is pre-trained on medical domain dialogue corpora. We also compared VRBot~\citep{vrbot}, which summarizes patient states and physician actions into phrases through variational method and generate the response.
In entity annotated dataset MedDG, we also compare with the entity concatenation method proposed by~\citet{meddg}, which predict the entities used in the response first, and then concatenate the predicted entities with history to produce the response.
Such two stages method has been verified to be effective in MedDG~\cite{meddg}. In the following, \emph{-Entity} suffix is used to distinguish the model with entity concatenation method.

\subsubsection{External Knowledge}
\label{sec:filter}
We exploit external knowledge following the previous knowledge-grounding dialogue generation methods~\citep{wizardofwiki, vrbot}, where the retrieved knowledge is encoded and fused in the decoder.
As verified by \citet{meddg}, predicting the medical entities used in the next response is helpful for informative response generation. Thus, we train our knowledge retrieval model to retrieve medical entities might be used in the response.

First, the medical entities appeared in the dialogue history are used as center nodes to select sub-graphs with one-hop relation in CMeKG. Then, we only retrieve entities contained in sub-graphs, which reduces the searching space for effective retrieval.
Inspired by the bi-encoder dense retrieval method \cite{dpr}, we employ two independent PCL-MedBERT to encodes dialogue history $X$ and any entity $E$ (consists of several tokens) respectively, and take the representation at the $[CLS]$ token as the encoder's output.

Denote the dialogue history encoding as $\mathbf{h}_{X}$, and the entity encoding as $\mathbf{h}_{E}$, the inner product of $\mathbf{h}_{X}$ and $\mathbf{h}_{E}$ denotes the score to retrieve this entity.
Let $E^+_i$ be one of the positive entity appeared in the target response, alone with $n$ negative entities $\{E^-_j\}_{j=1}^n$ not appeared. We optimize the loss function as the negative log likelihood of the positive entity:
\begin{equation}
\label{eq:dpr}
    \mathcal{L}_{X, E_i^+} = -\log \frac
    {
    \exp({\mathbf{h}_{X}^\mathsf{T} \mathbf{h}_{E^+_i}})
    }
    {
 \exp({\mathbf{h}_{X}^\mathsf{T} \mathbf{h}_{E^+_i}}) + \sum_{j=1}^n \exp({\mathbf{h}_{X}^\mathsf{T} \mathbf{h}_{E^-_j}})
    }
    .
\end{equation}
The losses produced by all positive entities in each example are averaged as the final loss to train the retriever.

We retrieve top $20$ entities for each dialogue history. This can be done with a single forward pass over datasets, where the retrieved entities are not changed during training and inference. Then, we employ an another PCL-MedBERT as the knowledge encoder to encode the retrieved entities. The retrieved entities are sorted by their retrieval scores and are concatenated by $[SEP]$ token to a sequence. The knowledge encoder encodes the sequence to knowledge encoding $\mathbf{H}_k$, and the encoder will be finetuned during training.

\subsubsection{Implementation Details}
\label{sec:implement}
For knowledge-aware graph encoder, the vertex embedding size and speaker embedding size is $512$, and we use $2$ layers RGAT~\cite{rgat} to encode the graph. For recall supervised signals construction, we set utterance number $k$ to $3$ in MedDG and $2$ in MedDialog, and constrained the recall utterances in the last six rounds of dialogue history. For the RNN-based models, the encoder and decoder consist of one layer LSTM. 
Both the size of word embedding and hidden states are set to $300$. 
For VRBot, we do not use the additional annotation of response intention for comparable experiments.
For pre-trained models BERT-GPT and DialoGPT, the configurations are following the original works. We use exploitable pre-trained parameters of BERT-GPT to initialize our model. Due to its decoder uses encoding from encoder through self-attention, we initialize the cross-attention modules from scratch. We also pre-trained our model on medical domain corpus that used in BERT-GPT to improve the performance. For entity prediction in MedDG, we use $10$-fold cross-validation models and ensemble results by majority voting method.
The learning rate is initialized to $10^{-4}$ and $10^{-5}$ for the RNN-based model and pre-trained model. The loss coefficients $\lambda_Y$ and $\lambda_\mathcal{R}$ are set to $0.9$, and $\lambda_r$ is set to $0.1$. 
We use the Adam optimizer~\citep{adam}, learning rate warm-up over the first $3000$ steps and linear decay of the learning rate.
Models generate responses through beam-sample\footnote{https://huggingface.co/transformers/internal/generation\_utils} algorithm, where beam-size and topk are set to $5$ and $64$. Other generation hyper-parameters keep default settings.
We use the NLTK toolkit with  \textit{SmoothFunction7} to calculate BLEU scores following~\citet{meddg}.

\subsection{Results and Analysis}

The automatic evaluation results are shown in \cref{tab:expresult} and \cref{tab:meddialog_exp}.
\ModelName{} outperforms other models both on BLEU and F1 metrics. 
As shown in Table \ref{tab:expresult}, BERT-GPT-\textit{Entity} is the model with the best all-around performance among comparative models.
Our model outperforms the strongest baseline model BERT-GPT-\textit{Entity} on BLEU-1/2/4 scores by a large margin, and outperforms it by $2$ points on F1. 
In addition, \ModelName{} outperforms BERT-GPT$^*$ by $3$ points on F1 and $1$ points on BLEU-1 (see \cref{tab:meddialog_exp}). 
These experimental results indicate that the proposed model is superior to previous models in terms of fluency and accuracy.
We can see that transformer-based models DialoGPT, BERT-GPT$^*$ and \ModelName{} performs significantly better than RNN-based models, e.g. DialoGPT outperforms VRBot by $4$ points on F1, suggesting the advantages of transformers-based models in larger dataset. 
Moreover, the experimental comparisons in DISTINCT-2 metric suggest our model reaches a competitive level in generating diverse responses when achieving new SOTA results on other evaluation metrics.

We also observe that all the models with \textit{-Entity} improves the BLEU-1 and F1 scores. It verifies the medical entities are useful knowledge for medical response generation.
But we also observe that the entity concatenation method is unstable, e.g., BERT-GPT-\emph{Entity} obtain worse BLEU-4 than BERT-GPT. 
It may be caused by the low medical entities prediction accuracy. 
In addition, it is costly to annotate the entities entailed in utterances. But it is necessary for the entity concatenate method.
By comparing the experimental results of MedPIR-\emph{KDGE \& REG} on F1 metric, we found that our knowledge retrieval strategy and gate-based fusion network are more effective and stable than other models. 

\begin{table}[h]
\renewcommand\arraystretch{1.2} 
\setlength\tabcolsep{4pt}
\centering
    \begin{tabular}{lcccc|c}
        \hline         \textbf{Model} & \textbf{B@1} & \textbf{B@2} & \textbf{B@4} & \textbf{D@2}  &  \textbf{F1} \\ \hline
        Seq2Seq~\citep{seq}         & 0.301 & 0.225 & 0.163 & 0.791 & 0.063    \\ \hline
        HRED~\citep{hred}           & 0.299 & 0.226 & 0.180 & 0.785 & 0.080    \\ \hline
        DialoGPT~\citep{dialogpt}        & 0.275 & 0.204 & 0.155 & 0.706 & 0.128    \\ \hline
        BERT-GPT$^*$~\citep{bertgpt}    & 0.298 & 0.232 & 0.202 & 0.821 & 0.145    \\ \hline
        VRBot~\citep{vrbot}         & 0.281 & 0.203 & 0.147 & 0.668 & 0.081    \\ \hline
        \hline
        \ModelName{} (Ours)      & \textbf{0.308} & \textbf{0.237} & \textbf{0.210} & 0.811 & \textbf{0.174} \\ \hline
        \hspace{0.4em}$-$ KDGE   
                & 0.291 & 0.229 & 0.201 & 0.825 & 0.158 \\ \hline
        \hspace{0.4em}$-$ REG
                        & 0.285 & 0.229 & 0.202 & 0.813 & 0.163
                      \\ \hline
        \hspace{0.4em}$-$ Knowledge encoder
                        & 0.296 & 0.231 & 0.202 & 0.817 & 0.164 \\ \hline
        \hspace{0.4em}$-$ KDGE \& REG
                        & 0.291 & 0.227 & 0.187 & \textbf{0.827} & 0.159
        \\ 
        \hline
    \end{tabular}
    \caption{Automatic evaluation results on MedDialog dataset. BERT-GPT$^*$ has been pre-trained on the MedDialog. REG indicates recall-enhanced generator, and KDGE indicates knowledge-enhanced dialogue graph encoder.}
    \label{tab:meddialog_exp}
\end{table}

\subsubsection{Ablation Study}
We also take the ablation experiments to verify the effects of different modules in \ModelName{}, which are presented in the last four lines of \cref{tab:expresult} and \cref{tab:meddialog_exp}.
The experimental results suggest both knowledge-aware dialogue graph encoder (KDGE) and recall-enhanced generator (REG) improve the medical response generation.
When we dropout the REG, where the generator produces responses directly, there is an obvious performance degradation on BLEU scores and a slight decrease on F1 score.
It suggests the effectiveness of training the model generates pivotal information weakly supervised by PCL-MedBERT.
When we only dropout the KDGE ($-$ KDGE), the performance decrease significantly on BLEU and F1 scores.
It indicates that the KDGE is vital to facilitate the recall-enhanced generator in \ModelName.
Though the model is trained to generate recall, there is only a modest improvement without structure encoding.
It is because the structure encoding captures the causal information from dialogue structure, supporting the model recalling long dialogue history effectively.
Finally, when we dropout KDGE \& REG, the performance decreases the most on BLUE metrics, 
indicating the effectiveness of the two main components in \ModelName.

As shown in \cref{tab:meddialog_exp}, the REG and KDGE improve less in MedDialog than in MedDG. We suggest that it may be attributed to the fact that the length of dialogue in MedDialog is relatively short, which is also pointed by~\citet{vrbot}. The average number of utterances in MedDialog ($9.5$, the version cleaned by~\citet{vrbot}) is less than MedDG ($21.5$). It shows that \ModelName{} could focus on pivotal information scattered in long dialogue history and has preferable performance as the conversation length increases.

\begin{figure}[b]
    \centering
    \includegraphics[scale=0.4]{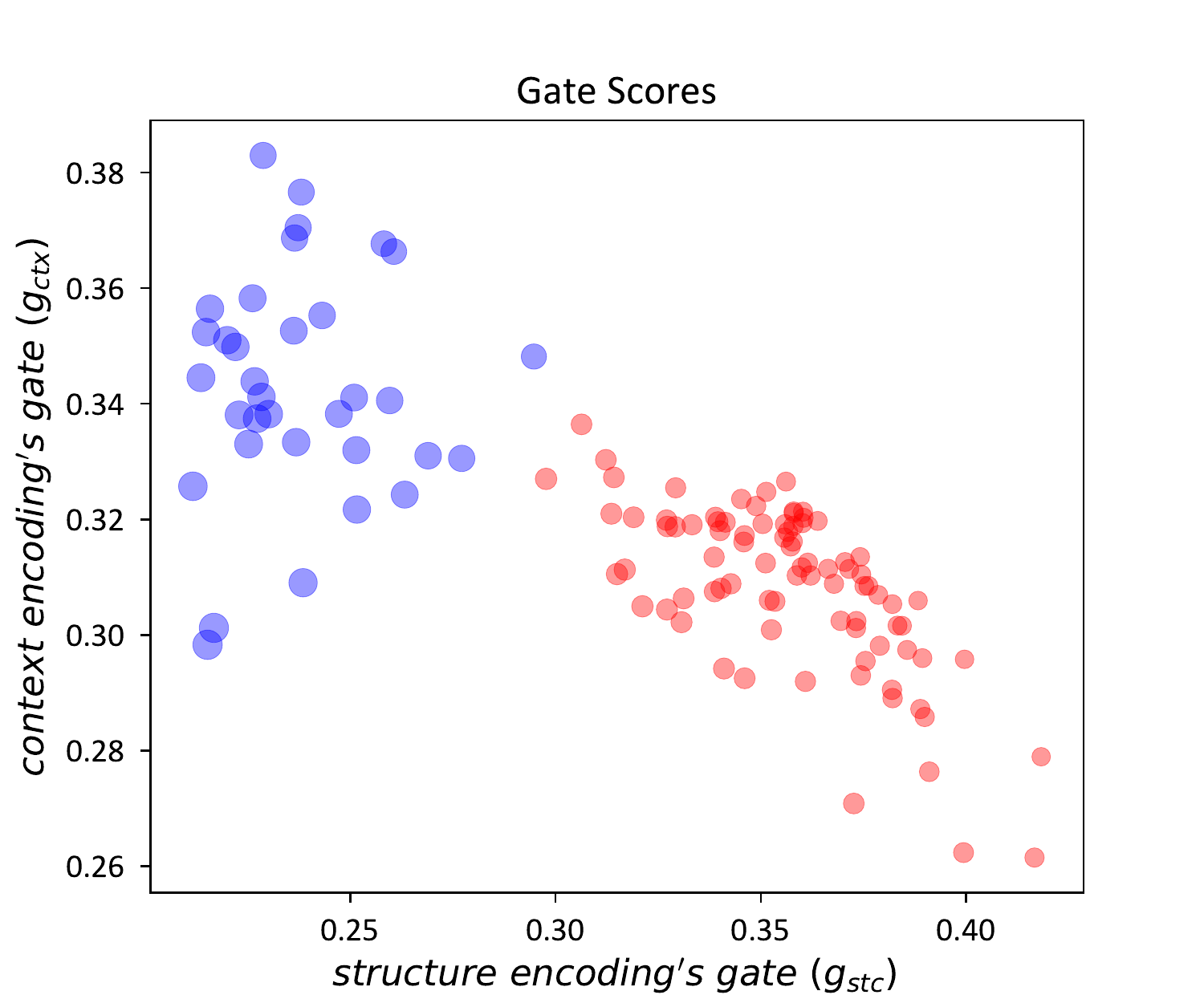}
    \caption{
    The blue dots and red dots represent tokens of response and recall respectively. The scale of the dot is proportional to the knowledge gate score.}
        \label{fig:gatescore}
\end{figure}
\subsubsection{Analysis of Multifaceted Encoding}
\label{sec:inflows}
We select an example from MedDG and draw the picture to show how the model uses dialogue structure encoding, context encoding and knowledge encoding during recalling pivotal information and generating response.
As shown in \cref{fig:gatescore}, the blue and red dots represent tokens of response and recall, respectively. The horizontal axis and vertical axis show the gates' scores $g_{stc}$ and $g_{ctx}$, respectively, and the scale of a dot is proportional to $g_{k}$. We observe that recall tokens distribute in the bottom-right part, and response tokens distribute in the upper-left part.
It indicates that the model mainly uses structure encoding when recalling pivotal information and mainly uses context encoding when generating the response.
This distribution shows that KDGE provides complementary information to the context encoding and facilitates REG to recall pivotal information.
Though the response generation uses less structure encoding, the generator can access the pre-generated recall sequence by self-attention.
The scales of blue dots are larger than red dots, indicating the model access knowledge information more when generating the response.

\subsubsection{Human Evaluation}
We conducted the human evaluation of responses in the aspects of fluency, consistency, and entity correctness.
We randomly selected 100 samples from the test set of MedDG, and the corresponding responses generated by well-performed models, e.g., DialoGPT, DialoGPT-\emph{Entity}, BERT-GPT, BERT-GPT-\emph{Entity} and \ModelName.
To ensure the fairness of assessment, the responses of each sample are shuffled and then provided to volunteers for evaluation. The final statistic results are shown in \cref{tab:humaneval}. Three manual evaluation indicators show that our proposed model still performs the best and far surpasses other models.
Especially in aspects of coherence and correctness, \ModelName{} significantly outperforms other compared models, suggesting that the proposed method improve the quality of responses. 

\begin{table}[b]
\renewcommand\arraystretch{1.2} 
\centering
\begin{tabular}{lccc}
\hline
\textbf{Model} & \textbf{Fluency} & \textbf{Coherence} & \textbf{Correctness} \\ \hline
DialoGPT        & 3.69 & 3.46 & 2.76 \\ \hline
DialoGPT-Entity & 4.30 & 3.20 & 2.84 \\ \hline
BERT-GPT        & 4.36 & 3.73 & 3.06 \\ \hline
BERT-GPT-Entity & 4.35 & 3.74 & 3.13 \\ \hline
\ModelName      & \textbf{4.42} & \textbf{3.86} & \textbf{3.25} \\ \hline
\end{tabular}
\caption{Results of human evaluation. The maximum score of each indicator is $5$. }
\label{tab:humaneval}
\end{table}

\begin{figure*}[!t]
    \footnotesize
    \centering
    \includegraphics[scale=0.75]{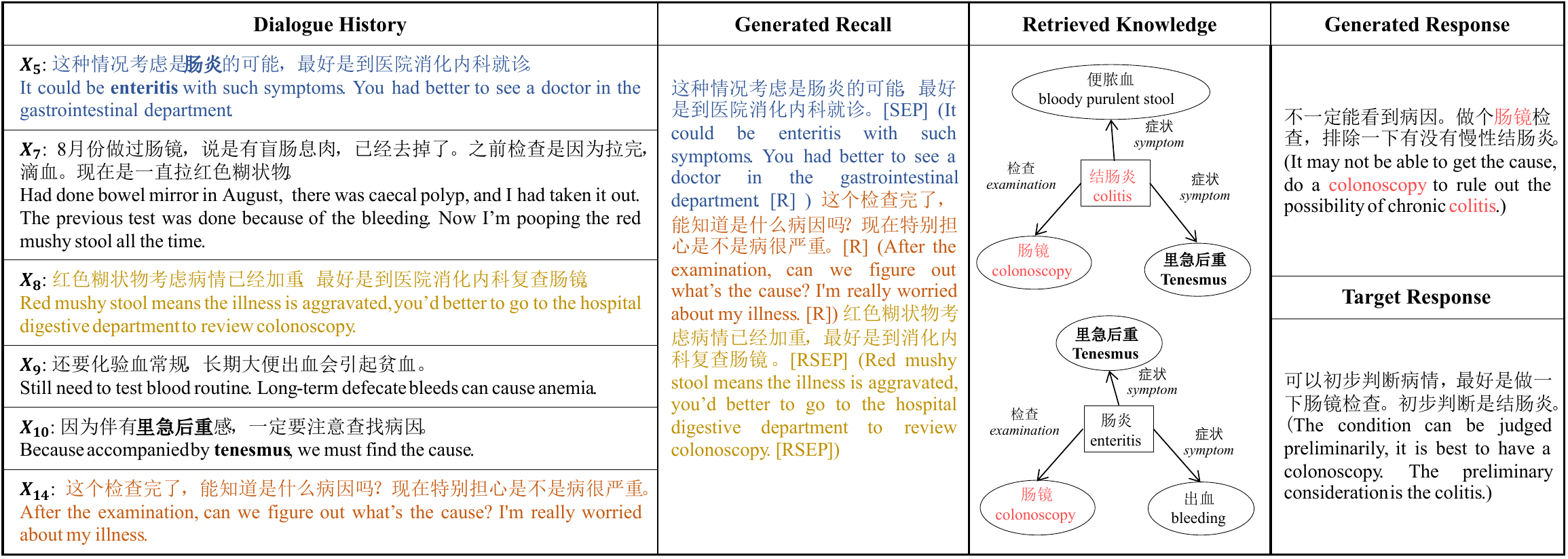}
    \caption{An example of recall and response generated by \ModelName{} in MedDG.
  The recalled utterances are colored correspondingly in the dialogue history. The bold entities in the history are used to retrieve knowledge. The retrieved knowledge with red-colored words are used in the generated response. }
    \label{fig:case}
\end{figure*}

\subsubsection{Case Study}
We present a case to show the pivotal information recalling method in our \ModelName~in Figure~\ref{fig:case}.
The model generates recalled utterances, including the history utterances $X_5, X_{14}$ and $X_{8}$ in order, which are colored correspondingly in the dialogue history.
The retrieved knowledge includes the symptoms and examinations about \textit{enteritis} and \textit{colitis} are colored by corresponding background colors in the dialogue history. 
\ModelName{} generates the responses conditioned on the retrieved knowledge and recalled utterances, which are presented in the second and third columns.
The generated response and target response are shown in the last column. We can observe that the generated response's semantics is similar to the target response, where both of them express that the patient may suffer the colitis and should do a colonoscopy.
The case indicates \ModelName~can generate responses with pivotal information recalling and use retrieved knowledge effectively.

\section{Conclusion}
In this paper, we propose a medical response generation model with pivotal information recalling~(\ModelName) to explicitly generate the pivotal information before producing the response.
In this way, the generator strengthens the interaction between the response and pivotal information from dialogue history. \ModelName{} mainly consists of a knowledge-aware dialogue graph encoder (KDGE) and a recall-enhanced generator (REG). KDGE constructs a dialogue graph by exploiting the knowledge relationships between entities in the utterances, and encodes the graph through a graph encoder. REG equipped with the gate module to incorporate multifaceted encodings, and it recalls the pivotal information and generates the response successively. Our experiments on MedDG and MedDialog datasets demonstrate the effectiveness of \ModelName.
\begin{acks}
We appreciate the insightful feedback from the anonymous reviewers. This work is jointly supported by grants: Natural Science Foundation of China (No. 62006061 and 61872107), Stable Support Program for Higher Education Institutions of Shenzhen (No. GXWD20201230155427003-20200824155011001) and Strategic Emerging Industry Development Special Funds of Shenzhen(No. JCYJ20200109113441941).
\end{acks}
\clearpage
\bibliographystyle{ACM-Reference-Format}

\begin{thebibliography}{34}


\ifx \showCODEN    \undefined \def \showCODEN     #1{\unskip}     \fi
\ifx \showDOI      \undefined \def \showDOI       #1{#1}\fi
\ifx \showISBNx    \undefined \def \showISBNx     #1{\unskip}     \fi
\ifx \showISBNxiii \undefined \def \showISBNxiii  #1{\unskip}     \fi
\ifx \showISSN     \undefined \def \showISSN      #1{\unskip}     \fi
\ifx \showLCCN     \undefined \def \showLCCN      #1{\unskip}     \fi
\ifx \shownote     \undefined \def \shownote      #1{#1}          \fi
\ifx \showarticletitle \undefined \def \showarticletitle #1{#1}   \fi
\ifx \showURL      \undefined \def \showURL       {\relax}        \fi
\providecommand\bibfield[2]{#2}
\providecommand\bibinfo[2]{#2}
\providecommand\natexlab[1]{#1}
\providecommand\showeprint[2][]{arXiv:#2}

\bibitem[Bahdanau et~al\mbox{.}(2015)]%
        {attention}
\bibfield{author}{\bibinfo{person}{Dzmitry Bahdanau},
  \bibinfo{person}{Kyunghyun Cho}, {and} \bibinfo{person}{Yoshua Bengio}.}
  \bibinfo{year}{2015}\natexlab{}.
\newblock \showarticletitle{Neural Machine Translation by Jointly Learning to
  Align and Translate}. In \bibinfo{booktitle}{\emph{3rd International
  Conference on Learning Representations, {ICLR} 2015, San Diego, CA, USA, May
  7-9, 2015, Conference Track Proceedings}},
  \bibfield{editor}{\bibinfo{person}{Yoshua Bengio} {and} \bibinfo{person}{Yann
  LeCun}} (Eds.).
\newblock


\bibitem[Busbridge et~al\mbox{.}(2019)]%
        {rgat}
\bibfield{author}{\bibinfo{person}{Dan Busbridge}, \bibinfo{person}{Dane
  Sherburn}, \bibinfo{person}{Pietro Cavallo}, {and} \bibinfo{person}{Nils~Y.
  Hammerla}.} \bibinfo{year}{2019}\natexlab{}.
\newblock \showarticletitle{Relational Graph Attention Networks}.
\newblock \bibinfo{journal}{\emph{CoRR}}  \bibinfo{volume}{abs/1904.05811}
  (\bibinfo{year}{2019}).
\newblock
\showeprint[arXiv]{1904.05811}


\bibitem[Byambasuren et~al\mbox{.}(2019)]%
        {cmekg}
\bibfield{author}{\bibinfo{person}{Odma Byambasuren}, \bibinfo{person}{Yunfei
  Yang}, \bibinfo{person}{Zhifang Sui}, \bibinfo{person}{Damai Dai},
  \bibinfo{person}{Baobao Chang}, \bibinfo{person}{Sujian Li}, {and}
  \bibinfo{person}{Hongying Zan}.} \bibinfo{year}{2019}\natexlab{}.
\newblock \showarticletitle{Preliminary study on the construction of Chinese
  medical knowledge graph}.
\newblock \bibinfo{journal}{\emph{Journal of Chinese Information Processing}}
  \bibinfo{volume}{33}, \bibinfo{number}{10} (\bibinfo{year}{2019}).
\newblock


\bibitem[Chen et~al\mbox{.}(2018)]%
        {chen-etal-2018-structured}
\bibfield{author}{\bibinfo{person}{Lu Chen}, \bibinfo{person}{Bowen Tan},
  \bibinfo{person}{Sishan Long}, {and} \bibinfo{person}{Kai Yu}.}
  \bibinfo{year}{2018}\natexlab{}.
\newblock \showarticletitle{Structured Dialogue Policy with Graph Neural
  Networks}. In \bibinfo{booktitle}{\emph{Proceedings of the 27th International
  Conference on Computational Linguistics, {COLING} 2018, Santa Fe, New Mexico,
  USA, August 20-26, 2018}}, \bibfield{editor}{\bibinfo{person}{Emily~M.
  Bender}, \bibinfo{person}{Leon Derczynski}, {and} \bibinfo{person}{Pierre
  Isabelle}} (Eds.). \bibinfo{publisher}{Association for Computational
  Linguistics}, \bibinfo{pages}{1257--1268}.
\newblock


\bibitem[Chen et~al\mbox{.}(2020a)]%
        {bertgpt}
\bibfield{author}{\bibinfo{person}{Shu Chen}, \bibinfo{person}{Zeqian Ju},
  \bibinfo{person}{Xiangyu Dong}, \bibinfo{person}{Hongchao Fang},
  \bibinfo{person}{Sicheng Wang}, \bibinfo{person}{Yue Yang},
  \bibinfo{person}{Jiaqi Zeng}, \bibinfo{person}{Ruisi Zhang},
  \bibinfo{person}{Ruoyu Zhang}, \bibinfo{person}{Meng Zhou},
  \bibinfo{person}{Penghui Zhu}, {and} \bibinfo{person}{Pengtao Xie}.}
  \bibinfo{year}{2020}\natexlab{a}.
\newblock \showarticletitle{MedDialog: {A} Large-scale Medical Dialogue
  Dataset}.
\newblock \bibinfo{journal}{\emph{CoRR}}  \bibinfo{volume}{abs/2004.03329}.
\newblock
\showeprint[arXiv]{2004.03329}


\bibitem[Chen et~al\mbox{.}(2020b)]%
        {chen2020bridging}
\bibfield{author}{\bibinfo{person}{Xiuyi Chen}, \bibinfo{person}{Fandong Meng},
  \bibinfo{person}{Peng Li}, \bibinfo{person}{Feilong Chen},
  \bibinfo{person}{Shuang Xu}, \bibinfo{person}{Bo Xu}, {and}
  \bibinfo{person}{Jie Zhou}.} \bibinfo{year}{2020}\natexlab{b}.
\newblock \showarticletitle{Bridging the Gap between Prior and Posterior
  Knowledge Selection for Knowledge-Grounded Dialogue Generation}. In
  \bibinfo{booktitle}{\emph{Proceedings of the 2020 Conference on Empirical
  Methods in Natural Language Processing, {EMNLP} 2020, Online, November 16-20,
  2020}}, \bibfield{editor}{\bibinfo{person}{Bonnie Webber},
  \bibinfo{person}{Trevor Cohn}, \bibinfo{person}{Yulan He}, {and}
  \bibinfo{person}{Yang Liu}} (Eds.). \bibinfo{publisher}{Association for
  Computational Linguistics}, \bibinfo{pages}{3426--3437}.
\newblock


\bibitem[Dinan et~al\mbox{.}(2019)]%
        {wizardofwiki}
\bibfield{author}{\bibinfo{person}{Emily Dinan}, \bibinfo{person}{Stephen
  Roller}, \bibinfo{person}{Kurt Shuster}, \bibinfo{person}{Angela Fan},
  \bibinfo{person}{Michael Auli}, {and} \bibinfo{person}{Jason Weston}.}
  \bibinfo{year}{2019}\natexlab{}.
\newblock \showarticletitle{Wizard of Wikipedia: Knowledge-Powered
  Conversational Agents}. In \bibinfo{booktitle}{\emph{7th International
  Conference on Learning Representations, {ICLR} 2019, New Orleans, LA, USA,
  May 6-9, 2019}}. \bibinfo{publisher}{OpenReview.net}.
\newblock


\bibitem[Du et~al\mbox{.}(2019a)]%
        {extract_symptoms}
\bibfield{author}{\bibinfo{person}{Nan Du}, \bibinfo{person}{Kai Chen},
  \bibinfo{person}{Anjuli Kannan}, \bibinfo{person}{Linh Tran},
  \bibinfo{person}{Yuhui Chen}, {and} \bibinfo{person}{Izhak Shafran}.}
  \bibinfo{year}{2019}\natexlab{a}.
\newblock \showarticletitle{Extracting Symptoms and their Status from Clinical
  Conversations}. In \bibinfo{booktitle}{\emph{Proceedings of the 57th
  Conference of the Association for Computational Linguistics, {ACL} 2019,
  Florence, Italy, July 28- August 2, 2019, Volume 1: Long Papers}},
  \bibfield{editor}{\bibinfo{person}{Anna Korhonen}, \bibinfo{person}{David~R.
  Traum}, {and} \bibinfo{person}{Llu{\'{\i}}s M{\`{a}}rquez}} (Eds.).
  \bibinfo{publisher}{Association for Computational Linguistics},
  \bibinfo{pages}{915--925}.
\newblock
\urldef\tempurl%
\url{https://doi.org/10.18653/v1/p19-1087}
\showDOI{\tempurl}


\bibitem[Du et~al\mbox{.}(2019b)]%
        {learn-to-infer-entity}
\bibfield{author}{\bibinfo{person}{Nan Du}, \bibinfo{person}{Mingqiu Wang},
  \bibinfo{person}{Linh Tran}, \bibinfo{person}{Gang Lee}, {and}
  \bibinfo{person}{Izhak Shafran}.} \bibinfo{year}{2019}\natexlab{b}.
\newblock \showarticletitle{Learning to Infer Entities, Properties and their
  Relations from Clinical Conversations}. In
  \bibinfo{booktitle}{\emph{Proceedings of the 2019 Conference on Empirical
  Methods in Natural Language Processing and the 9th International Joint
  Conference on Natural Language Processing (EMNLP-IJCNLP)}}.
  \bibinfo{pages}{4979--4990}.
\newblock


\bibitem[Feng et~al\mbox{.}(2021)]%
        {ddams}
\bibfield{author}{\bibinfo{person}{Xiachong Feng}, \bibinfo{person}{Xiaocheng
  Feng}, \bibinfo{person}{Bing Qin}, {and} \bibinfo{person}{Xinwei Geng}.}
  \bibinfo{year}{2021}\natexlab{}.
\newblock \showarticletitle{Dialogue Discourse-Aware Graph Model and Data
  Augmentation for Meeting Summarization}. In
  \bibinfo{booktitle}{\emph{Proceedings of the Thirtieth International Joint
  Conference on Artificial Intelligence, {IJCAI} 2021, Virtual Event /
  Montreal, Canada, 19-27 August 2021}},
  \bibfield{editor}{\bibinfo{person}{Zhi{-}Hua Zhou}} (Ed.).
  \bibinfo{publisher}{ijcai.org}, \bibinfo{pages}{3808--3814}.
\newblock
\urldef\tempurl%
\url{https://doi.org/10.24963/ijcai.2021/524}
\showDOI{\tempurl}


\bibitem[Ghazvininejad et~al\mbox{.}(2018)]%
        {ghazvininejad2018knowledge}
\bibfield{author}{\bibinfo{person}{Marjan Ghazvininejad},
  \bibinfo{person}{Chris Brockett}, \bibinfo{person}{Ming-Wei Chang},
  \bibinfo{person}{Bill Dolan}, \bibinfo{person}{Jianfeng Gao},
  \bibinfo{person}{Wen-tau Yih}, {and} \bibinfo{person}{Michel Galley}.}
  \bibinfo{year}{2018}\natexlab{}.
\newblock \showarticletitle{A knowledge-grounded neural conversation model}. In
  \bibinfo{booktitle}{\emph{Proceedings of the AAAI Conference on Artificial
  Intelligence}}, Vol.~\bibinfo{volume}{32}.
\newblock


\bibitem[Karpukhin et~al\mbox{.}(2020)]%
        {dpr}
\bibfield{author}{\bibinfo{person}{Vladimir Karpukhin}, \bibinfo{person}{Barlas
  Oguz}, \bibinfo{person}{Sewon Min}, \bibinfo{person}{Patrick Lewis},
  \bibinfo{person}{Ledell Wu}, \bibinfo{person}{Sergey Edunov},
  \bibinfo{person}{Danqi Chen}, {and} \bibinfo{person}{Wen-tau Yih}.}
  \bibinfo{year}{2020}\natexlab{}.
\newblock \showarticletitle{Dense Passage Retrieval for Open-Domain Question
  Answering}. In \bibinfo{booktitle}{\emph{Proceedings of the 2020 Conference
  on Empirical Methods in Natural Language Processing (EMNLP)}}.
  \bibinfo{pages}{6769--6781}.
\newblock


\bibitem[Kim et~al\mbox{.}(2020)]%
        {sequential-kg}
\bibfield{author}{\bibinfo{person}{Byeongchang Kim}, \bibinfo{person}{Jaewoo
  Ahn}, {and} \bibinfo{person}{Gunhee Kim}.} \bibinfo{year}{2020}\natexlab{}.
\newblock \showarticletitle{Sequential Latent Knowledge Selection for
  Knowledge-Grounded Dialogue}. In \bibinfo{booktitle}{\emph{8th International
  Conference on Learning Representations, {ICLR} 2020, Addis Ababa, Ethiopia,
  April 26-30, 2020}}. \bibinfo{publisher}{OpenReview.net}.
\newblock


\bibitem[Kingma and Ba(2015)]%
        {adam}
\bibfield{author}{\bibinfo{person}{Diederik~P. Kingma} {and}
  \bibinfo{person}{Jimmy Ba}.} \bibinfo{year}{2015}\natexlab{}.
\newblock \showarticletitle{Adam: {A} Method for Stochastic Optimization}. In
  \bibinfo{booktitle}{\emph{3rd International Conference on Learning
  Representations, {ICLR} 2015, San Diego, CA, USA, May 7-9, 2015, Conference
  Track Proceedings}}, \bibfield{editor}{\bibinfo{person}{Yoshua Bengio} {and}
  \bibinfo{person}{Yann LeCun}} (Eds.).
\newblock


\bibitem[Li et~al\mbox{.}(2021)]%
        {vrbot}
\bibfield{author}{\bibinfo{person}{Dongdong Li}, \bibinfo{person}{Zhaochun
  Ren}, \bibinfo{person}{Pengjie Ren}, \bibinfo{person}{Zhumin Chen},
  \bibinfo{person}{Miao Fan}, \bibinfo{person}{Jun Ma}, {and}
  \bibinfo{person}{Maarten de Rijke}.} \bibinfo{year}{2021}\natexlab{}.
\newblock \showarticletitle{Semi-Supervised Variational Reasoning for Medical
  Dialogue Generation}. In \bibinfo{booktitle}{\emph{{SIGIR} '21: The 44th
  International {ACM} {SIGIR} Conference on Research and Development in
  Information Retrieval, Virtual Event, Canada, July 11-15, 2021}},
  \bibfield{editor}{\bibinfo{person}{Fernando Diaz}, \bibinfo{person}{Chirag
  Shah}, \bibinfo{person}{Torsten Suel}, \bibinfo{person}{Pablo Castells},
  \bibinfo{person}{Rosie Jones}, {and} \bibinfo{person}{Tetsuya Sakai}} (Eds.).
  \bibinfo{publisher}{{ACM}}, \bibinfo{pages}{544--554}.
\newblock
\urldef\tempurl%
\url{https://doi.org/10.1145/3404835.3462921}
\showDOI{\tempurl}


\bibitem[Li et~al\mbox{.}(2016)]%
        {distinct}
\bibfield{author}{\bibinfo{person}{Jiwei Li}, \bibinfo{person}{Michel Galley},
  \bibinfo{person}{Chris Brockett}, \bibinfo{person}{Jianfeng Gao}, {and}
  \bibinfo{person}{William~B Dolan}.} \bibinfo{year}{2016}\natexlab{}.
\newblock \showarticletitle{A Diversity-Promoting Objective Function for Neural
  Conversation Models}. In \bibinfo{booktitle}{\emph{Proceedings of the 2016
  Conference of the North American Chapter of the Association for Computational
  Linguistics: Human Language Technologies}}. \bibinfo{pages}{110--119}.
\newblock


\bibitem[Li et~al\mbox{.}(2020)]%
        {li2020zero}
\bibfield{author}{\bibinfo{person}{Linxiao Li}, \bibinfo{person}{Can Xu},
  \bibinfo{person}{Wei Wu}, \bibinfo{person}{Yufan Zhao},
  \bibinfo{person}{Xueliang Zhao}, {and} \bibinfo{person}{Chongyang Tao}.}
  \bibinfo{year}{2020}\natexlab{}.
\newblock \showarticletitle{Zero-resource knowledge-grounded dialogue
  generation}.
\newblock \bibinfo{journal}{\emph{Advances in Neural Information Processing
  Systems}}  \bibinfo{volume}{33} (\bibinfo{year}{2020}),
  \bibinfo{pages}{8475--8485}.
\newblock


\bibitem[Lian et~al\mbox{.}(2019)]%
        {learningtoselect}
\bibfield{author}{\bibinfo{person}{Rongzhong Lian}, \bibinfo{person}{Min Xie},
  \bibinfo{person}{Fan Wang}, \bibinfo{person}{Jinhua Peng}, {and}
  \bibinfo{person}{Hua Wu}.} \bibinfo{year}{2019}\natexlab{}.
\newblock \showarticletitle{Learning to Select Knowledge for Response
  Generation in Dialog Systems}. In \bibinfo{booktitle}{\emph{Proceedings of
  the Twenty-Eighth International Joint Conference on Artificial Intelligence,
  {IJCAI} 2019, Macao, China, August 10-16, 2019}},
  \bibfield{editor}{\bibinfo{person}{Sarit Kraus}} (Ed.).
  \bibinfo{publisher}{ijcai.org}, \bibinfo{pages}{5081--5087}.
\newblock


\bibitem[Lin et~al\mbox{.}(2021)]%
        {graph-evolving}
\bibfield{author}{\bibinfo{person}{Shuai Lin}, \bibinfo{person}{Pan Zhou},
  \bibinfo{person}{Xiaodan Liang}, \bibinfo{person}{Jianheng Tang},
  \bibinfo{person}{Ruihui Zhao}, \bibinfo{person}{Ziliang Chen}, {and}
  \bibinfo{person}{Liang Lin}.} \bibinfo{year}{2021}\natexlab{}.
\newblock \showarticletitle{Graph-Evolving Meta-Learning for Low-Resource
  Medical Dialogue Generation}. In \bibinfo{booktitle}{\emph{Thirty-Fifth
  {AAAI} Conference on Artificial Intelligence, {AAAI} 2021, Thirty-Third
  Conference on Innovative Applications of Artificial Intelligence, {IAAI}
  2021, The Eleventh Symposium on Educational Advances in Artificial
  Intelligence, {EAAI} 2021, Virtual Event, February 2-9, 2021}}.
  \bibinfo{publisher}{{AAAI} Press}.
\newblock


\bibitem[Lin et~al\mbox{.}(2019)]%
        {enhance-graph}
\bibfield{author}{\bibinfo{person}{Xinzhu Lin}, \bibinfo{person}{Xiahui He},
  \bibinfo{person}{Qin Chen}, \bibinfo{person}{Huaixiao Tou},
  \bibinfo{person}{Zhongyu Wei}, {and} \bibinfo{person}{Ting Chen}.}
  \bibinfo{year}{2019}\natexlab{}.
\newblock \showarticletitle{Enhancing dialogue symptom diagnosis with global
  attention and symptom graph}. In \bibinfo{booktitle}{\emph{Proceedings of the
  2019 Conference on Empirical Methods in Natural Language Processing and the
  9th International Joint Conference on Natural Language Processing
  (EMNLP-IJCNLP)}}. \bibinfo{pages}{5033--5042}.
\newblock


\bibitem[Liu et~al\mbox{.}(2020)]%
        {meddg}
\bibfield{author}{\bibinfo{person}{Wenge Liu}, \bibinfo{person}{Jianheng Tang},
  \bibinfo{person}{Jinghui Qin}, \bibinfo{person}{Lin Xu},
  \bibinfo{person}{Zhen Li}, {and} \bibinfo{person}{Xiaodan Liang}.}
  \bibinfo{year}{2020}\natexlab{}.
\newblock \showarticletitle{MedDG: {A} Large-scale Medical Consultation Dataset
  for Building Medical Dialogue System}.
\newblock \bibinfo{journal}{\emph{CoRR}}  \bibinfo{volume}{abs/2010.07497}
  (\bibinfo{year}{2020}).
\newblock
\showeprint[arXiv]{2010.07497}


\bibitem[Madotto et~al\mbox{.}(2020)]%
        {madotto2020learning}
\bibfield{author}{\bibinfo{person}{Andrea Madotto}, \bibinfo{person}{Samuel
  Cahyawijaya}, \bibinfo{person}{Genta~Indra Winata}, \bibinfo{person}{Yan Xu},
  \bibinfo{person}{Zihan Liu}, \bibinfo{person}{Zhaojiang Lin}, {and}
  \bibinfo{person}{Pascale Fung}.} \bibinfo{year}{2020}\natexlab{}.
\newblock \showarticletitle{Learning Knowledge Bases with Parameters for
  Task-Oriented Dialogue Systems}. In \bibinfo{booktitle}{\emph{Findings of the
  Association for Computational Linguistics: {EMNLP} 2020, Online Event, 16-20
  November 2020}} \emph{(\bibinfo{series}{Findings of {ACL}},
  Vol.~\bibinfo{volume}{{EMNLP} 2020})},
  \bibfield{editor}{\bibinfo{person}{Trevor Cohn}, \bibinfo{person}{Yulan He},
  {and} \bibinfo{person}{Yang Liu}} (Eds.). \bibinfo{publisher}{Association for
  Computational Linguistics}, \bibinfo{pages}{2372--2394}.
\newblock


\bibitem[Papineni et~al\mbox{.}(2002)]%
        {bleu}
\bibfield{author}{\bibinfo{person}{Kishore Papineni}, \bibinfo{person}{Salim
  Roukos}, \bibinfo{person}{Todd Ward}, {and} \bibinfo{person}{Wei{-}Jing
  Zhu}.} \bibinfo{year}{2002}\natexlab{}.
\newblock \showarticletitle{Bleu: a Method for Automatic Evaluation of Machine
  Translation}. In \bibinfo{booktitle}{\emph{Proceedings of the 40th Annual
  Meeting of the Association for Computational Linguistics, July 6-12, 2002,
  Philadelphia, PA, {USA}}}. \bibinfo{publisher}{{ACL}},
  \bibinfo{pages}{311--318}.
\newblock


\bibitem[See et~al\mbox{.}(2017)]%
        {pinternet}
\bibfield{author}{\bibinfo{person}{Abigail See}, \bibinfo{person}{Peter~J.
  Liu}, {and} \bibinfo{person}{Christopher~D. Manning}.}
  \bibinfo{year}{2017}\natexlab{}.
\newblock \showarticletitle{Get To The Point: Summarization with
  Pointer-Generator Networks}. In \bibinfo{booktitle}{\emph{Proceedings of the
  55th Annual Meeting of the Association for Computational Linguistics, {ACL}
  2017, Vancouver, Canada, July 30 - August 4, Volume 1: Long Papers}},
  \bibfield{editor}{\bibinfo{person}{Regina Barzilay} {and}
  \bibinfo{person}{Min{-}Yen Kan}} (Eds.). \bibinfo{publisher}{Association for
  Computational Linguistics}, \bibinfo{pages}{1073--1083}.
\newblock


\bibitem[Serban et~al\mbox{.}(2016)]%
        {hred}
\bibfield{author}{\bibinfo{person}{Iulian~Vlad Serban},
  \bibinfo{person}{Alessandro Sordoni}, \bibinfo{person}{Yoshua Bengio},
  \bibinfo{person}{Aaron~C. Courville}, {and} \bibinfo{person}{Joelle Pineau}.}
  \bibinfo{year}{2016}\natexlab{}.
\newblock \showarticletitle{Building End-To-End Dialogue Systems Using
  Generative Hierarchical Neural Network Models}. In
  \bibinfo{booktitle}{\emph{Proceedings of the Thirtieth {AAAI} Conference on
  Artificial Intelligence, February 12-17, 2016, Phoenix, Arizona, {USA}}},
  \bibfield{editor}{\bibinfo{person}{Dale Schuurmans} {and}
  \bibinfo{person}{Michael~P. Wellman}} (Eds.). \bibinfo{publisher}{{AAAI}
  Press}, \bibinfo{pages}{3776--3784}.
\newblock


\bibitem[Song et~al\mbox{.}(2020)]%
        {med_dialogue_smmary}
\bibfield{author}{\bibinfo{person}{Yan Song}, \bibinfo{person}{Yuanhe Tian},
  \bibinfo{person}{Nan Wang}, {and} \bibinfo{person}{Fei Xia}.}
  \bibinfo{year}{2020}\natexlab{}.
\newblock \showarticletitle{Summarizing Medical Conversations via Identifying
  Important Utterances}. In \bibinfo{booktitle}{\emph{Proceedings of the 28th
  International Conference on Computational Linguistics, {COLING} 2020,
  Barcelona, Spain (Online), December 8-13, 2020}},
  \bibfield{editor}{\bibinfo{person}{Donia Scott}, \bibinfo{person}{N{\'{u}}ria
  Bel}, {and} \bibinfo{person}{Chengqing Zong}} (Eds.).
  \bibinfo{publisher}{International Committee on Computational Linguistics},
  \bibinfo{pages}{717--729}.
\newblock


\bibitem[Speer et~al\mbox{.}(2017)]%
        {conceptnet}
\bibfield{author}{\bibinfo{person}{Robyn Speer}, \bibinfo{person}{Joshua Chin},
  {and} \bibinfo{person}{Catherine Havasi}.} \bibinfo{year}{2017}\natexlab{}.
\newblock \showarticletitle{ConceptNet 5.5: An Open Multilingual Graph of
  General Knowledge}. In \bibinfo{booktitle}{\emph{Proceedings of the
  Thirty-First {AAAI} Conference on Artificial Intelligence, February 4-9,
  2017, San Francisco, California, {USA}}},
  \bibfield{editor}{\bibinfo{person}{Satinder Singh} {and}
  \bibinfo{person}{Shaul Markovitch}} (Eds.). \bibinfo{publisher}{{AAAI}
  Press}, \bibinfo{pages}{4444--4451}.
\newblock


\bibitem[Sun et~al\mbox{.}(2021)]%
        {graph}
\bibfield{author}{\bibinfo{person}{Yajing Sun}, \bibinfo{person}{Yong Shan},
  \bibinfo{person}{Chengguang Tang}, \bibinfo{person}{Yue Hu},
  \bibinfo{person}{Yinpei Dai}, \bibinfo{person}{Jing Yu},
  \bibinfo{person}{Jian Sun}, \bibinfo{person}{Fei Huang}, {and}
  \bibinfo{person}{Luo Si}.} \bibinfo{year}{2021}\natexlab{}.
\newblock \showarticletitle{Unsupervised Learning of Deterministic Dialogue
  Structure with Edge-Enhanced Graph Auto-Encoder}. In
  \bibinfo{booktitle}{\emph{Proceedings of the AAAI Conference on Artificial
  Intelligence}}, Vol.~\bibinfo{volume}{35}. \bibinfo{pages}{13869--13877}.
\newblock


\bibitem[Sutskever et~al\mbox{.}(2014)]%
        {seq}
\bibfield{author}{\bibinfo{person}{Ilya Sutskever}, \bibinfo{person}{Oriol
  Vinyals}, {and} \bibinfo{person}{Quoc~V. Le}.}
  \bibinfo{year}{2014}\natexlab{}.
\newblock \showarticletitle{Sequence to Sequence Learning with Neural
  Networks}. In \bibinfo{booktitle}{\emph{Advances in Neural Information
  Processing Systems 27: Annual Conference on Neural Information Processing
  Systems 2014, December 8-13 2014, Montreal, Quebec, Canada}},
  \bibfield{editor}{\bibinfo{person}{Zoubin Ghahramani}, \bibinfo{person}{Max
  Welling}, \bibinfo{person}{Corinna Cortes}, \bibinfo{person}{Neil~D.
  Lawrence}, {and} \bibinfo{person}{Kilian~Q. Weinberger}} (Eds.).
  \bibinfo{pages}{3104--3112}.
\newblock


\bibitem[Vaswani et~al\mbox{.}(2017)]%
        {transformer}
\bibfield{author}{\bibinfo{person}{Ashish Vaswani}, \bibinfo{person}{Noam
  Shazeer}, \bibinfo{person}{Niki Parmar}, \bibinfo{person}{Jakob Uszkoreit},
  \bibinfo{person}{Llion Jones}, \bibinfo{person}{Aidan~N. Gomez},
  \bibinfo{person}{Lukasz Kaiser}, {and} \bibinfo{person}{Illia Polosukhin}.}
  \bibinfo{year}{2017}\natexlab{}.
\newblock \showarticletitle{Attention is All you Need}. In
  \bibinfo{booktitle}{\emph{Advances in Neural Information Processing Systems
  30: Annual Conference on Neural Information Processing Systems 2017, December
  4-9, 2017, Long Beach, CA, {USA}}},
  \bibfield{editor}{\bibinfo{person}{Isabelle Guyon}, \bibinfo{person}{Ulrike
  von Luxburg}, \bibinfo{person}{Samy Bengio}, \bibinfo{person}{Hanna~M.
  Wallach}, \bibinfo{person}{Rob Fergus}, \bibinfo{person}{S.~V.~N.
  Vishwanathan}, {and} \bibinfo{person}{Roman Garnett}} (Eds.).
  \bibinfo{pages}{5998--6008}.
\newblock


\bibitem[Xu et~al\mbox{.}(2021)]%
        {xu-etal-2021-discovering}
\bibfield{author}{\bibinfo{person}{Jun Xu}, \bibinfo{person}{Zeyang Lei},
  \bibinfo{person}{Haifeng Wang}, \bibinfo{person}{Zheng-Yu Niu},
  \bibinfo{person}{Hua Wu}, {and} \bibinfo{person}{Wanxiang Che}.}
  \bibinfo{year}{2021}\natexlab{}.
\newblock \showarticletitle{Discovering Dialog Structure Graph for Coherent
  Dialog Generation}. In \bibinfo{booktitle}{\emph{Proceedings of the 59th
  Annual Meeting of the Association for Computational Linguistics and the 11th
  International Joint Conference on Natural Language Processing (Volume 1: Long
  Papers)}}. \bibinfo{pages}{1726--1739}.
\newblock


\bibitem[Zhang et~al\mbox{.}(2020a)]%
        {mie}
\bibfield{author}{\bibinfo{person}{Yuanzhe Zhang}, \bibinfo{person}{Zhongtao
  Jiang}, \bibinfo{person}{Tao Zhang}, \bibinfo{person}{Shiwan Liu},
  \bibinfo{person}{Jiarun Cao}, \bibinfo{person}{Kang Liu},
  \bibinfo{person}{Shengping Liu}, {and} \bibinfo{person}{Jun Zhao}.}
  \bibinfo{year}{2020}\natexlab{a}.
\newblock \showarticletitle{MIE: A medical information extractor towards
  medical dialogues}. In \bibinfo{booktitle}{\emph{Proceedings of the 58th
  Annual Meeting of the Association for Computational Linguistics}}.
  \bibinfo{pages}{6460--6469}.
\newblock


\bibitem[Zhang et~al\mbox{.}(2020b)]%
        {dialogpt}
\bibfield{author}{\bibinfo{person}{Yizhe Zhang}, \bibinfo{person}{Siqi Sun},
  \bibinfo{person}{Michel Galley}, \bibinfo{person}{Yen-Chun Chen},
  \bibinfo{person}{Chris Brockett}, \bibinfo{person}{Xiang Gao},
  \bibinfo{person}{Jianfeng Gao}, \bibinfo{person}{Jingjing Liu}, {and}
  \bibinfo{person}{William~B Dolan}.} \bibinfo{year}{2020}\natexlab{b}.
\newblock \showarticletitle{DIALOGPT: Large-Scale Generative Pre-training for
  Conversational Response Generation}. In \bibinfo{booktitle}{\emph{Proceedings
  of the 58th Annual Meeting of the Association for Computational Linguistics:
  System Demonstrations}}. \bibinfo{pages}{270--278}.
\newblock


\bibitem[Zhao et~al\mbox{.}(2020)]%
        {low-resource-kg}
\bibfield{author}{\bibinfo{person}{Xueliang Zhao}, \bibinfo{person}{Wei Wu},
  \bibinfo{person}{Chongyang Tao}, \bibinfo{person}{Can Xu},
  \bibinfo{person}{Dongyan Zhao}, {and} \bibinfo{person}{Rui Yan}.}
  \bibinfo{year}{2020}\natexlab{}.
\newblock \showarticletitle{Low-Resource Knowledge-Grounded Dialogue
  Generation}. In \bibinfo{booktitle}{\emph{8th International Conference on
  Learning Representations, {ICLR} 2020, Addis Ababa, Ethiopia, April 26-30,
  2020}}. \bibinfo{publisher}{OpenReview.net}.
\newblock


\end{thebibliography}

\end{document}